\documentclass{article}

\usepackage{microtype}
\usepackage{graphicx}
\usepackage{subfigure}
\usepackage{booktabs} %

\usepackage{hyperref}

\usepackage[accepted]{icml2021}

\usepackage{enumitem}
\setlist{nosep}

\icmltitlerunning{An Empirical Investigation of Learning from Biased Toxicity Labels}

\setlist[description,1]{leftmargin=0.7em,labelindent=*}

\usepackage[compact]{titlesec}
\titlespacing{\section}{0pt}{1ex}{0ex}
\titlespacing{\subsection}{0pt}{0ex}{0ex}
\begin{document}

\twocolumn[
\icmltitle{An Empirical Investigation of Learning from Biased Toxicity Labels}

\icmlsetsymbol{equal}{*}
\icmlsetsymbol{intern}{*}

\begin{icmlauthorlist}
\icmlauthor{Neel Nanda}{dm,intern}
\icmlauthor{Sven Gowal}{dm}
\icmlauthor{Jonathan Uesato}{dm}
\end{icmlauthorlist}

\icmlaffiliation{dm}{DeepMind, London, UK}

\icmlkeywords{Machine Learning, ICML}

\icmlcorrespondingauthor{Neel Nanda}{neelnanda27@gmail.com}

\vskip 0.3in
]

\printAffiliationsAndNotice{\textsuperscript{*}Work completed during an internship at DeepMind.\linebreak}%

\begin{abstract}
Collecting annotations from human raters often results in a trade-off between the quantity of labels one wishes to gather and the quality of these labels. As such, it is often only possible to gather a small amount of high-quality labels. In this paper, we study how different training strategies can leverage a small dataset of human-annotated labels and a large but noisy dataset of synthetically generated labels (which exhibit bias against identity groups) for predicting toxicity of online comments. We evaluate the accuracy and fairness properties of these approaches, 
and trade-offs between the two.
While we find that initial training on all of the data and fine-tuning on clean data produces models with the highest AUC,
we find that no single strategy performs best across all fairness metrics.
\end{abstract}
\vspace{-1em}
\section{Introduction}
\label{intro}

Supervised learning systems rely on labeled training data, often human-annotated. This creates a trade-off. Human raters are imperfect and introduce bias and variance into their labels \cite{geva-etal-2019-modeling}. When given enough time and resources, the quality of such labels can improve dramatically \cite{stiennon2020learning}, resulting in labels with lower bias and variance. Hence, given a fixed budget, there is a trade-off between label quality and quantity. One possible solution to this trade-off is to create a large amount of cheap, low-quality labels and a small amount of expensive, high-quality labels. This enables novel training approaches that use high-quality labels to minimise biases learnt from low-quality labels \cite{xiao2015learning,ren2018learning, zhang2020distilling,song2020learning}.

Natural language is a particularly compelling context to study, as the field has seen recent rapid progress \cite{BERT, GPT3} and models are becoming increasingly widely deployed, yet often exhibit bias \citep{dixon2018measuring, kurita2019measuring, sap2019risk}. In this work, we explore different ways to train a fair textual toxicity classifier \citep{wulczyn2017ex, dixon2018measuring, CivilComments} in this regime. We have access to a small amount of human-generated high-quality labels and a large amount of synthetically-generated low-quality labels. Our low-quality labels exhibit fairness-relevant biases, in particular, systematic differences in accuracy and predicted toxicity rate for different identity groups. Our high-quality labels also exhibit these biases, but to a lesser degree.

We formalise this as a noisy labels problem, where we have a dataset of noisy labels (low-quality) and of clean labels (high-quality). To study this problem, we build a setup with the following key properties:

\begin{description}
\item[Labeler type] The training data is annotated with labeler type - we know whether each data point is clean or noisy, and we have data of each type. This is in contrast to work that assumes we can only train on the noisy data \cite{Jiang2019}.
\item[Imbalance] We have significantly more noisy data than clean data. Class imbalance is known to introduce bias and undesirable learning dynamics in other settings \cite{yang2020rethinking}.
\item[Complex bias] The biases are difficult to model precisely and often qualitative, as they emerge from human judgement. This is in contrast to prior work that models noisy labels as flipping labels between classes according to a transition matrix, independent of the input \cite{Hendrycks2018,lamy2019noise}.
\end{description}

We focus on the Civil Comments dataset \cite{CivilComments}, a collection of online comments annotated as toxic or non-toxic. 
This is suitable for a study of fairness as comments are annotated by identity references, enabling measurement of unintended bias against protected groups. 

Similarly to \citet{NoisyLabels}, we synthetically generate noisy labels, as there is no preexisting clean/noisy label split in the Civil Comments dataset. We treat the human labels as clean and train toxicity classification models on the human labels to generate synthetic labels. We explore several standard approaches to training models from imperfect data. We evaluate their accuracy and bias, and whether there is a trade-off between the two. 

We find that initial training on all of the data followed by fine-tuning on the clean data is the best way to train an accurate model. Measuring fairness is more complex, and the right approach depends on the specific context where a model will be applied \citep{barocas2017fairness}. Accordingly, we use the fairness-relevant metrics introduced by \citet{CivilComments} to evaluate a range of possible biases. We focus on metrics that measure systematic differences in accuracy and systematic differences in predicted toxicity rate for different demographic groups. We find that no single model performs best on all metrics.

\section{Methods}
\label{sec:experiment}

\subsection{Data}
\label{sec:experiment setup}

For our investigation, we create a modified version of the Civil Comments dataset, a collection of almost two million online comments labeled as toxic and non-toxic. Our modified version consists of a combination of noisy (biased) and clean (less biased) labeled examples. To do so, we follow the approach of \citet{NoisyLabels} and take the existing human annotated labels in Civil Comments as our clean labels, and use predictions from a neural network classifier as our noisy labels. We now describe this data generation procedure in more detail.

The original human labels are our clean labels and to create the noisy labels, we train networks to imitate these human-annotated labels. We then use these synthetic raters to generate synthetic labels for each comment, our noisy labels. To ensure a suitable level of noise, we stop training the networks before convergence, attaining a validation set AUC of 95\%. See Section 3.3 for comparison with noisy labels generated by synthetic raters that are trained for longer. To avoid the synthetic labels being memorised from the training data, we hold out half of the dataset when training the synthetic raters and only generate synthetic labels for the held-out portion.

We create our clean and noisy datasets for training our baselines from the held-out portion. We re-label 95\% with noisy labels (i.e. discard the original clean label for that subset of data points), and the other 5\% retains the original human label. Our noisy dataset contains 878,620 examples and our clean dataset contains 46,232 examples, creating the desired imbalance between clean and noisy dataset size.

\textbf{Synthetic Label Biases} \quad Prior work has shown that networks trained on Civil Comments develop biases for or against identity groups, where different groups have systematic differences in accuracy and predicted toxicity rates \cite{dixon2018measuring, CivilComments}. This is attributed in part to correlations in the dataset such that comments mentioning certain identity groups are more or less likely to be toxic, and models tend to exaggerate this bias \cite{CivilComments}. Thus our noisy labels exhibit bias relative to the human labels, as required for our analysis. While the original human labels may also exhibit bias, we refer to them as clean to indicate that they are \textit{less} biased, not that they are unbiased.

\textbf{Limitations} \quad Naturally, this approach has the key limitation that our noisy labels are synthetically generated, rather than being generated by true human labelers. We are limited by the lack of publicly available datasets with well-defined clean/noisy splits, and which allow us to measure fairness properties. As argued in \citet{NoisyLabels}, we consider our approach a useful simulation of human bias. Neural network errors are complex and difficult to model, and share similarities with human error that simpler synthetic methods miss, such as having a higher error rate on harder examples.

See Appendix A for a more detailed discussion of how we generate this data, the properties of our noisy dataset and the limitations of this approach.

\subsection{Baselines}

We train several baselines on this synthetic dataset. All baselines are based on a pre-trained BERT \cite{BERT} encoder, followed by a 2-layer MLP. When training, we update the weights of both BERT and the MLP. All data points are of the form $(k, x, y)$, where $x$ is the comment text, the labeler type $k\in \{C, N\}$ represents whether the label is clean or noisy, and $y$ is the comment label. We train the baselines on both the clean and noisy datasets, and validate on clean data only.

We evaluate the following strategies:
\begin{description}
    \item[Clean] The model only trains on the clean data (5\% of the total training data).
    \item[Naive] The model trains on both clean and noisy data, and ignores the labeler type.
    \item[Multi-head] The model has two heads, and uses one for clean data points, one for noisy data points. Parameters in all prior layers are shared.
    \item[One-hot] The labeler type is one-hot encoded and appended to the BERT output before entering the MLP. 
    \item[Loss correction] \citep{LossCorrection} The noisy data is modelled as a corrupted version of the clean data, where for each pair of classes there is a certain fixed probability that each element of the first is corrupted to the second. The parameters of the corruption matrix are estimated from the available clean data, and applied to the model outputs when predicting noisy labels. No corruption is applied when predicting clean labels.
\end{description}

We further fine-tune each baseline (except \textit{clean}) on clean data. We denote this by appending the suffix \textbf{FT} to name of the baseline.

See Appendix B for further discussion of the technical specifications and behaviour of each baseline.

\section{Experiments}
\label{sec:Results}
\subsection{Accuracy}

We first measure the performance of each baseline, as measured by Area Under the ROC Curve (AUC) with respect to the clean labels. The AUC over the course of training for each baseline can be seen in Figure \ref{fig:Full AUC}, showing initial training on both clean and noisy data, and then fine-tuning on just clean data. This is calculated as AUC on the validation set, which has only clean labels. We observe that fine-tuning on clean data performs best (with a final AUC of 94.9\%), then \textit{multi-head} (with 94.7\%), and then \textit{clean}, \textit{naive} and \textit{one-hot} all perform similarly (between 94.2\% and 94.3\%). Notably, after fine-tuning on clean data all methods obtain similar performance (between 94.86\% and 94.94\%) despite significant variation in performance before fine-tuning. While we primarily use AUC to measure classification performance due to significant class imbalance, we note that our reported ordering between baselines is robust to alternative metrics such as binary accuracy and cross-entropy loss.

\subsection{Fairness}

\textbf{Metrics} \quad To measure fairness we use the fairness-relevant metrics introduced by \citet{CivilComments}, a common method for measuring bias in textual toxicity classification tasks \cite{civilcomments_kaggle,nozza2019unintended,zorian2019debiasing}. The Civil Comments Identities dataset is a subset of Civil Comments with annotations for whether each comment is a member of 13 identity groups, covering a range of race, religion, sexuality and gender considerations, allowing us to evaluate these metrics for each identity group. In particular, we focus on three of the metrics:

\begin{description}
    \item[Subgroup AUC] Evaluate the AUC of the model on each subgroup.
    \item[Background Positive, Subgroup Negative AUC] (BPSN AUC) Evaluate the AUC of the model on the non-toxic data points of the subgroup and toxic data points not of the subgroup.
    \item[Negative Average Equality Gap] (Negative AEG) Randomly select a non-toxic data point from the subgroup and a non-toxic data point not of the subgroup. Evaluate the proportion of the time that the model's predicted toxicity is higher for the subgroup data point. We subtract $0.5$, so that an unbiased model has $0$ Negative AEG.
\end{description}

We chose to focus on these metrics as they measure two distinct types of bias in our noisy data. Subgroup AUC measures systematic differences in accuracy between different identity groups. In contrast, Negative AEG measures differences in predicted toxicity rates, detecting cases where the model is more likely to label comments as toxic if they reference particular identity groups. BPSN AUC measures a combination of the two \cite{CivilComments}.

\begin{figure}[t]
\centering
\includegraphics[width=\columnwidth]{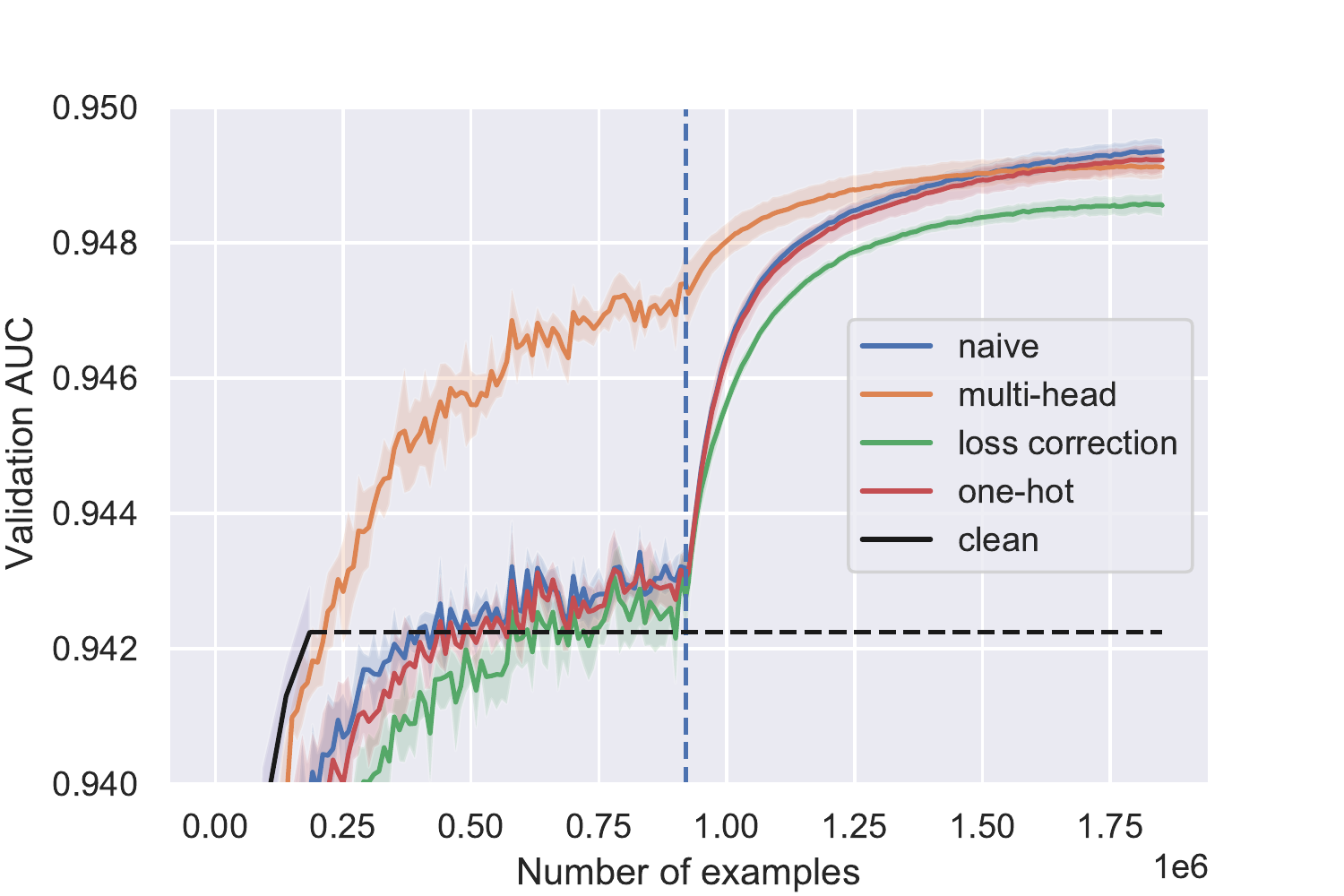}
\caption{AUC for each baseline. The vertical line is the start of training on clean data only. Before this, \textit{multi-head} has highest AUC. After training on clean data only, all baselines improve and AUC difference become smaller.}
\label{fig:Full AUC}
\end{figure}

We distinguish between \textbf{accuracy-based} metrics which correlate with overall AUC, and \textbf{accuracy-agnostic} metrics which do not. Subgroup AUC and BPSN AUC are accuracy-based as they measure model AUC on subsets of the data. Negative AEG is accuracy-agnostic, as a uniformly random classifier has a perfect Negative AEG of $0$.

\begin{figure*}[h]
\centering
\includegraphics[width=\linewidth]{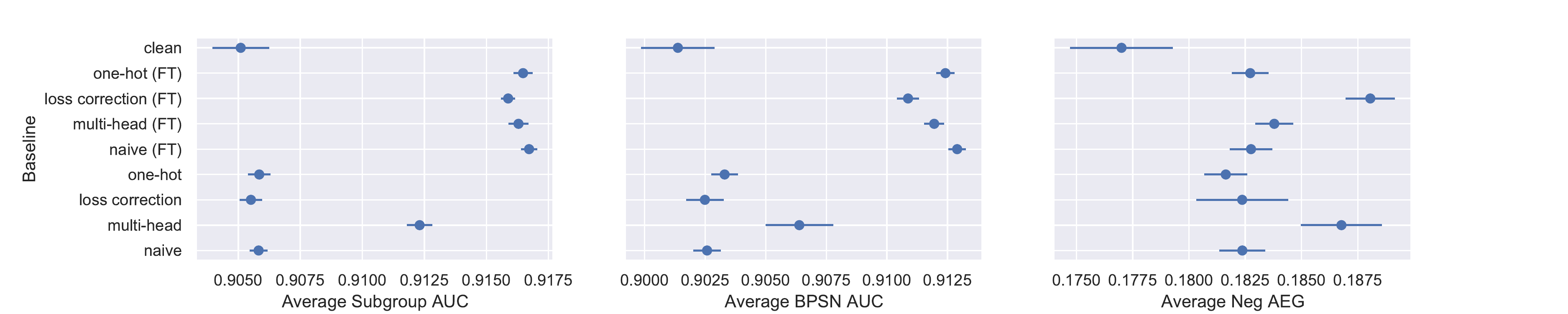}
\caption{The Subgroup AUC, Background Positive Subgroup Negative AUC (BPSN AUC) and Negative Average Equality Gap (Neg AEG) for each baseline, averaged over the 13 identity groups. Each baseline was run 5 times with different seeds, and the mean and standard deviation of the aggregated metric are plotted. Low Subgroup AUC and BPSN AUC and high Neg AEG indicate bias.}
\label{fig:Old bias}
\end{figure*}

\textbf{Results} \quad We measure the Subgroup AUC, BPSN AUC and Negative AEG for each baseline, for each of 13 identity groups. The results are displayed in Figure \ref{fig:Old bias}. We aggregate the metrics across the 13 identity groups by taking the arithmetic mean. Alternate approaches such as weighting by identity group size give similar results, and ordering for each metric is consistent across most subgroups, with a few exceptions. 

For the Subgroup AUC and BPSN AUC metrics, the fine-tuned baselines exhibit least bias, followed by \textit{multi-head}. However, this is the same ordering as overall AUC, as shown in Figure \ref{fig:Full AUC}. As these metrics are accuracy-based and correlate with overall AUC, it is difficult to determine whether this effect is due to lower bias, or a consequence of higher overall AUC.

\begin{figure}[t]
\centering
\includegraphics[width=\columnwidth]{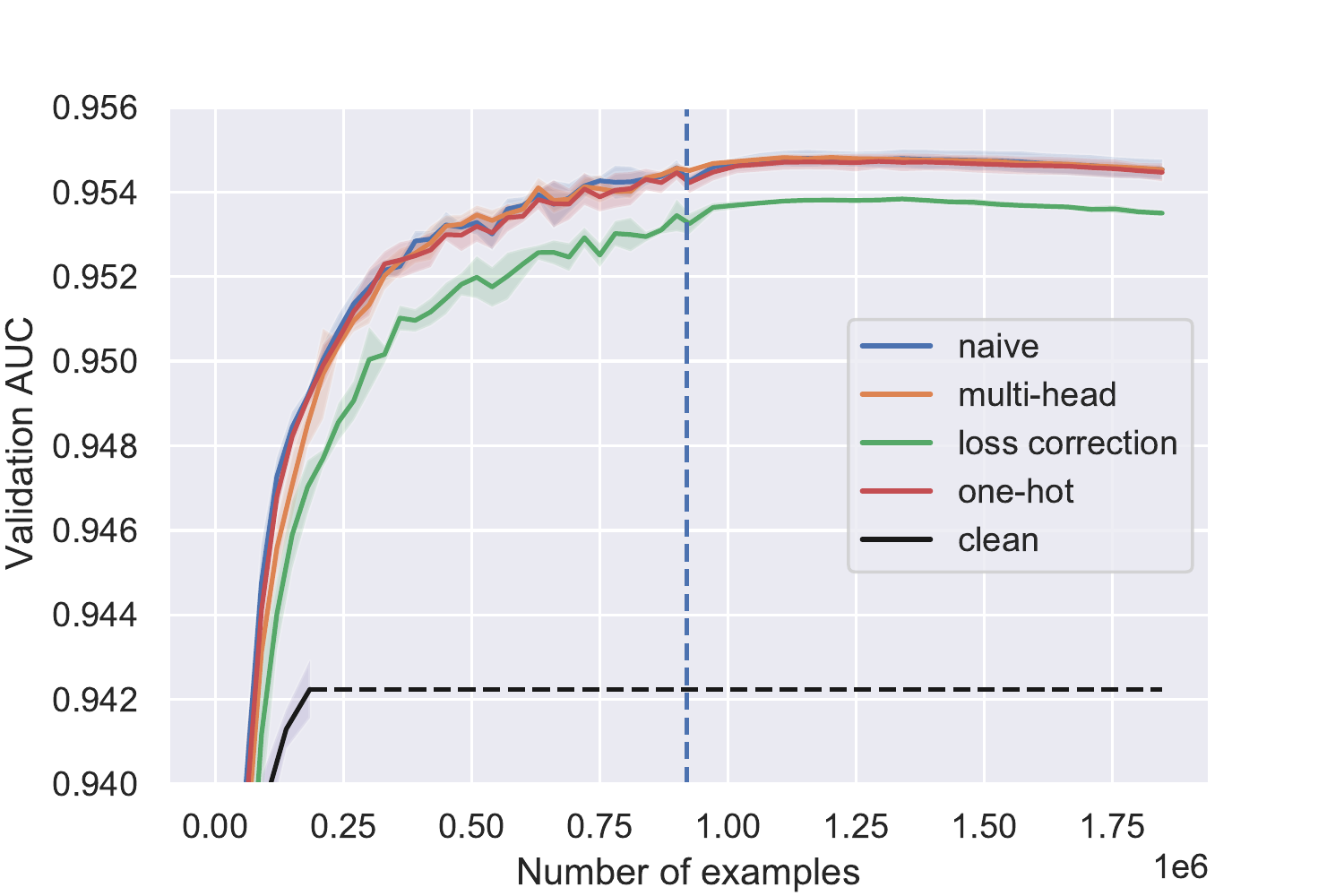}
\caption{Baseline AUC on higher-quality noisy data. The vertical dashed line shows the switch to training on clean data. Each baseline was run with 5 different seeds.}
\label{fig:New Performance}
\end{figure}

For the Negative AEG metric, the clean baseline exhibits the least bias and we find that all algorithms leveraging the noisy data introduce similar levels of bias. However, the difference in bias, though statistically significant, is slight and all baselines exhibit notable bias. The probability of classifying a subgroup data point as more toxic ranges between 67.7\% and 68.2\% for all models, significantly above the 50\% which would be attained by an unbiased classifier. 

Negative AEG is an accuracy-agnostic metric, meaning it just detects differences in bias, while Subgroup AUC and BPSN AUC are accuracy-based and detect differences in bias \textit{and} accuracy. Baselines fine-tuned on clean data perform best in terms of accuracy, Subgroup AUC and BPSN AUC yet not in terms of Negative AEG, and Negative AEG finds small overall differences in bias. This suggests the performance of fine-tuning may be attributed to higher overall AUC rather than decreased bias.

\subsection{Robustness Checks}

To investigate the robustness of our results, we explore the sensitivity to the level of noise. We generate a new synthetic dataset of noisy labels of better quality, reproduce our baselines and measurements and compare the results to our results for the original noisy labels. We produce higher-quality noisy labels by training our synthetic raters on more data points than before: 880,000 data points with batch size 16, in comparison to 220,000 data points with batch size 4. The accuracy of the noisy labels relative to the clean labels increases from 95\% to 95.5\% and the AUC increases from 96\% to 97\%.

Overall, orderings are preserved for both accuracy and fairness, but the magnitude of performance differences across baselines is significantly decreased. AUC for each baseline on the higher quality noisy labels is shown in Figure \ref{fig:New Performance}. Fine-tuning on clean data has highest overall AUC, though is an improvement on \textit{naive} of 0.1\%, in comparison to an improvement of 0.6\% before. \textit{Multi-head} AUC is similar to \textit{naive} and \textit{one-hot} before fine-tuning. The orderings for fairness metrics are similar to before: the accuracy-based metrics, Subgroup AUC and BPSN AUC, have the same ordering as overall AUC and \textit{clean} remains least biased under Negative AEG. We discuss this experiment further in Appendix C.

It is clear that performance will depend on the degree of difference between clean and noisy data. However, this was an unexpected level of sensitivity and suggests that the results of this paper may be fairly context-specific.

\section{Conclusion}
\label{sec:Takeaways}

In this work, we conducted an empirical investigation into learning from biased labels for toxicity prediction, using synthetic labels from a neural network as a proxy for noisy human labels.
With respect to AUC, models fine-tuned on clean data performed best on our original dataset.
With respect to fairness metrics, no single model performed best for all metrics -- while models fine-tuned on clean data exhibited the least bias on the accuracy-based metrics of Subgroup AUC and BPSN AUC, the approach of ignoring noisy labels entirely exhibited the least bias on the accuracy-agnostic Negative AEG metric.

As training machine learning models on large amounts of loosely curated data becomes commonplace, it is essential that we understand the effects of imperfect labels on accuracy and fairness. 
We recommend caution in extrapolating to other contexts based on these results -- we only study a single model architecture trained on a single dataset with synthetically generated labels, and different comparisons may result from different noise characteristics.
Nevertheless, we hope this work provides a useful set of empirical observations towards this important question.

\section*{Acknowledgements}

We thank Keren Gu for helpful comments and for creating initial versions of the synthetic data, Katerina Tsihlas for project management and general inspiration and support, and Johannes Welbl for helpful comments and feedback on earlier versions of this paper.

\bibliography{imperfect_labelers}
\bibliographystyle{icml2021}

\appendix
\clearpage
\newpage

\section{Discussion of the clean and noisy data}
\label{appendixa}

\begin{figure}[t]
\centering
\includegraphics[width=\columnwidth]{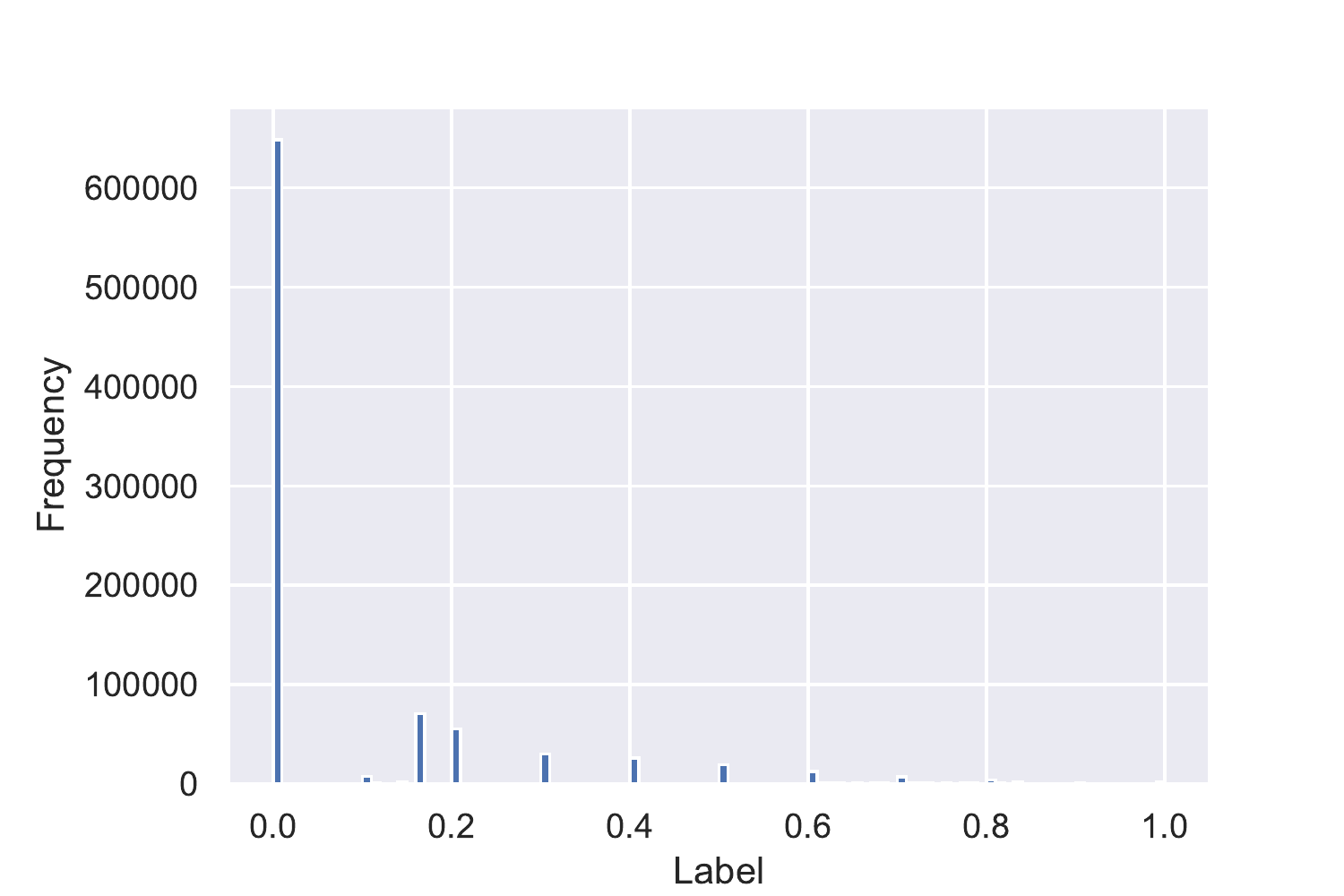}
\caption{Distribution of the clean labels}
\label{fig:Clean Hist}
\end{figure}

The original Civil Comments dataset was generated by having multiple human raters vote for whether a comment is toxic or non-toxic, and then was labeled with the proportion of votes for toxic. Most comments have 4 or 6 raters, while a few have significantly more raters. This means the data has soft labels, but that most labels take on a few common values. See Figure \ref{fig:Clean Hist} for the distribution. Another key feature of the dataset is the significant class imbalance. A substantial majority of comments had all raters vote non-toxic, and very few had all raters vote toxic. 

To generate our synthetic labels, we trained 9 models as synthetic raters. Each model used a BERT encoder followed by a linear classifier, trained to minimize the cross-entropy loss with respect to the soft labels, following \citet{wulczyn2017ex}. We trained models for each learning rate in $[10^{-5}, 3\times 10^{-6}, 10^{-6}]$ with 3 random seeds each. They were trained on 220,000 comments with batch size 4 and generated synthetic labels for a disjoint set of 920,000 comments. To ensure an appropriate degree of label noise, we train the raters for a single epoch, stopping before convergence.

\begin{figure}[t]
\centering
\includegraphics[width=\columnwidth]{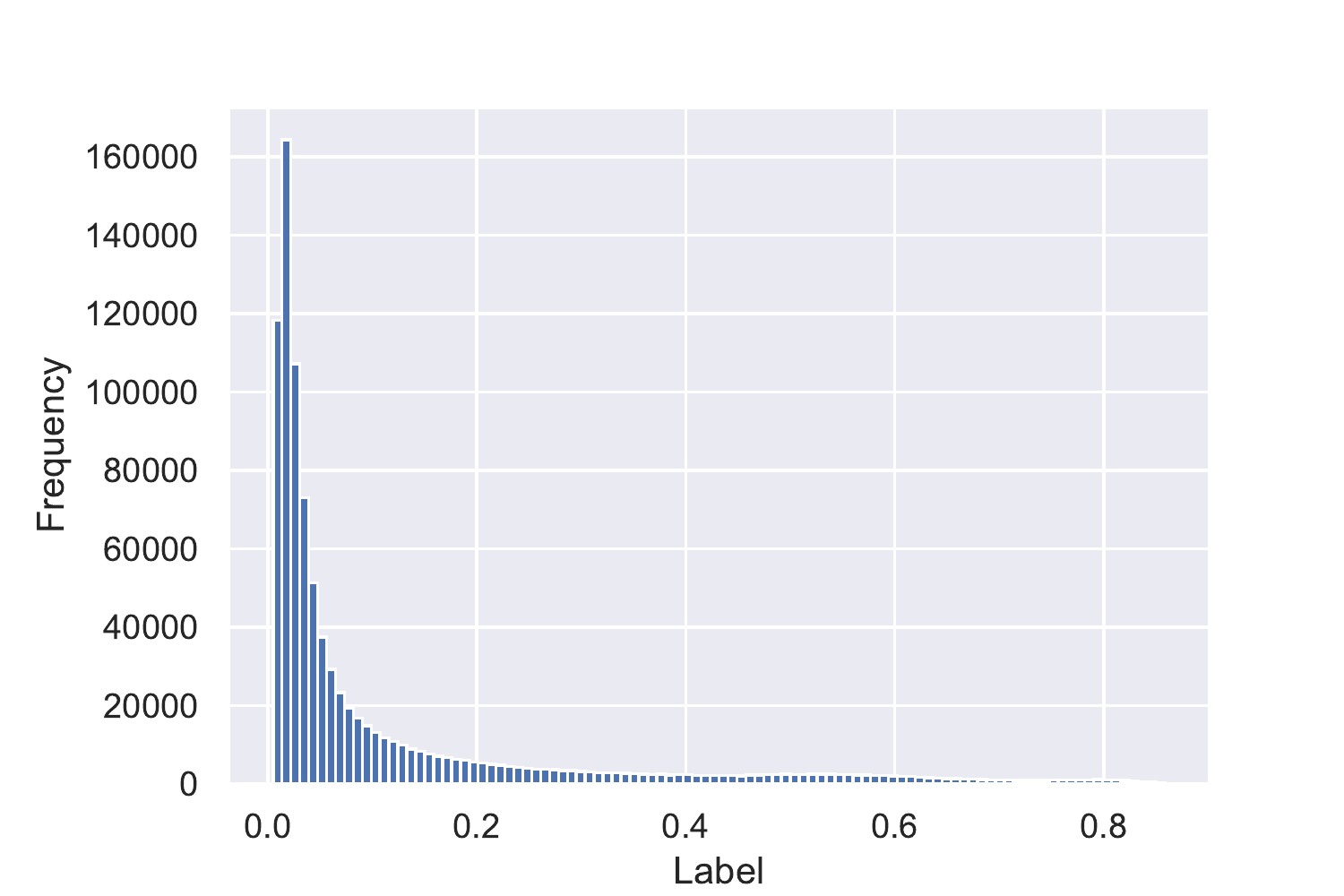}
\caption{Distribution of the noisy labels}
\label{fig:Noisy Hist}
\end{figure}

To calculate the synthetic labels, we average the 9 probabilities output by our synthetic raters. Ideally we would simulate the original process exactly, round each probability to 0 or 1, take this as a 'vote' by each model, and average. But the models are highly correlated, and this gives unacceptable accuracy hits, so we consider the deviation from the original procedure to be acceptable. See Figure \ref{fig:Noisy Hist} for the resulting distribution.

The synthetic data is broadly well calibrated at predicting the clean data, except at the tails, where the clean labels are significantly more likely to take on extreme values. The synthetic raters are trained with cross-entropy loss where outputting an extreme value incorrectly is heavily penalised, while outputting a correct extreme value is only somewhat rewarding. This was exacerbated by averaging the 9 models, since an extreme average value requires multiple models to output extreme values. As a result, though the noisy labels were in general calibrated, they were poorly calibrated at the tails. See Table \ref{tab:Quantiles} for details.

\begin{table*}[t]
\caption{Quantiles for clean and noisy data.}
 \vspace{0.9em}
 \begin{tabular}{l|llllllllll}
 
 \toprule
 Quantile &  0.000 &  0.001 &  0.010 &  0.020 &  0.100 &  0.900 &  0.980 &  0.990 &  0.999 &  1.000 \\
 \midrule
 Clean & 0.0000 & 0.0000 & 0.0000 & 0.0000 & 0.0000 & 0.4000 & 0.7600 & 0.8333 & 1.0000 & 1.0000 \\
 Noisy & 0.0051 & 0.0064 & 0.0081 & 0.0089 & 0.0125 & 0.3844 & 0.6855 & 0.7652 & 0.8394 & 0.8585 \\
 \bottomrule
\end{tabular}
\label{tab:Quantiles}

\end{table*}

\section{Technical Details of Baselines}
\label{appendixb}

We append a 2 layer MLP of layer sizes $768\to 768 \to 2$ to the pooled output of BERT. Each comment is entered as the first input to BERT, and we leave the second input empty.

Each model (other than clean) has an initial phase of being trained on the synthetic data (95\% noisy, 5\% clean) for 1 epoch and learning rate $3\times 10^{-6}$, and then fine-tuned on the clean data for 20 epochs with learning rate $10^{-7}$. The clean baseline is just trained on the clean data for up to 20 epochs with learning rate $3 \times 10^{-6}$. We used a batch size of 16.

The following are various side observations about each baseline. We note that these findings are less carefully documented than our results in other parts of our paper. Thus, while we caution drawing any definite conclusions from these observations, we include them to provide loose empirical intuition for the task to interested readers:

\vspace{-2ex}
\paragraph{Clean} This baseline starts to overfit fairly rapidly. We applied early stopping, stopping at the epoch with highest validation AUC. In practice, this was in epochs 2 to 5
\vspace{-2ex}
\paragraph{Multi-head} This baseline trains less stably than the rest. This is likely due to the clean head only being used for 1/20 data points, so a batch is often purely noisy data. The noisy head is trained end-to-end and as a result the representation in the penultimate layer sometimes shifts in line with the noisy head, but the clean head has a lag before it adjusts to the new representation. At batch size 4 this baseline was extremely unstable and noisy, at batch size 16 it exhibits less instability.
\vspace{-2ex}
\paragraph{One-hot} This baseline performs very similarly to \textit{naive}. It is possible as the labeler type is only appended at the start of the MLP, the model is simply not expressive enough to substantially account for this
\vspace{-2ex}
\paragraph{Loss correction} The setting here was quite different from the original paper \cite{LossCorrection}, and we had to make some adaptions. We took the approach the paper used for Clothing-1M, another dataset with clean and noisy data. We estimated the corruption probabilities by finding sets of perfect examples of each class (toxic and non-toxic) and looking at the average of the noisy labels. We applied the forward method from the paper, where to calculate loss we applied the corruption matrix to the network probabilities before calculating cross-entropy loss on the labels. We trained on the masked data, and applied the corruption for noisy labels, not for clean labels.
\begin{itemize}
\item As the data has soft labels, based on the votes of multiple human annotators, we define a perfect example of a non-toxic comment as one where all human annotators voted non-toxic. Due to significant class imbalance, very few comments had all human annotators vote toxic, so to get a more reliable estimate we define perfect examples of toxic comments to be where 90\% of the annotators voted toxic.
\item Perfect examples need to have both clean and noisy labels, which our existing setup does not allow. We relax the setup by generating synthetic noisy labels for our clean data points. To minimise the deviation from the experimental setup, loss correction \textit{only} uses these additional noisy labels to estimate corruption matrix parameters.
\item The resulting matrix had a 5\% of corruption for non-toxic to toxic and 40\% for toxic to non-toxic. We expect the 40\% is still a fairly noisy estimate.
\item As there were only two classes, with probabilities $p$ and $1-p$, the effect of the corruption matrix was essentially to apply a linear transformation to $p$, the probability output by the network. Note that the linear transformation was applied to the probabilities, not the logits. This significantly changed the calibration of the network, as outputting 0\% for a noisy comment was mapped to 5\%. As there were significant class imbalances and most comments were non-toxic, this incentivised the network to output probabilities much closer to 0, as cross-entropy loss heavily penalises an incorrect answer of 0\% relative to an incorrect answer of 5\%. This problem was significantly worse if just trained on noisy data, rather than 95\% noisy and 5\% clean, as the corruption was not applied for clean data, so outputting 0\% was still majorly penalised.
\end{itemize}

\section{Investigating Sensitivity to Levels of Noise}

We reproduce our experiments with the same setup but higher-quality noisy data. Here we train synthetic raters with 880,000 data points and batch size 16, rather than 220,000 data points with batch size 4. Correspondingly, AUC for the synthetic labelers increases from 96\% to 97\%, and accuracy increases from 95\% to 95.5\%. 

The AUC of baselines trained on this data can be seen in Figure \ref{fig:New Performance}. Note that \textit{clean} is identical to before, as the clean data is kept the same. The most striking difference is that \textit{multi-head} now does exactly as well as \textit{one-hot} and \textit{naive}. Fine-tuning on clean data is still an improvement over \textit{naive}, but represents a much smaller improvement and starts overfitting. As a result, we apply early stopping to both \textit{clean} and all fine-tuning steps. 

\begin{figure*}[b]
\centering
\includegraphics[width=\linewidth]{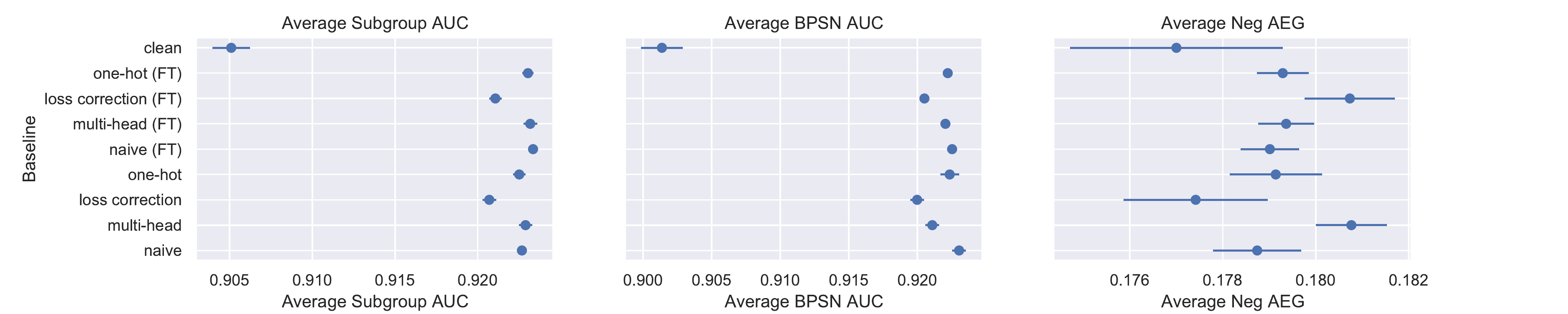}
\caption{The Subgroup AUC, Background Positive Subgroup Negative AUC (BPSN AUC) and Negative Average Equality Gap (Neg AEG) for each baseline when trained on higher quality noisy data, averaged over the 13 identity groups. Each baseline was run 5 times with different seeds, and the mean and standard deviation of the aggregated metric are plotted. Low Subgroup AUC and BPSN AUC and high Neg AEG indicate bias.}
\label{fig:New bias}
\end{figure*}

Fairness results are shown in Figure \ref{fig:New bias}. These demonstrate a similar pattern to before. On the accuracy-based metrics of Subgroup AUC and BPSN AUC, the ranking is similar to the ranking of accuracy. For the accuracy-agnostic metric of Negative AEG, \textit{clean} remains the best, though by a smaller margin, as the noisy data exhibits less bias. In all cases, we early stop in fine-tuning based on validation AUC.

It is clear that our results will be somewhat sensitive to the level of label noise, e.g. if the noisy labels are the same quality as clean labels, then the naive baseline should perform best. But we find it striking how such a seemingly small increase in quality has such a significant difference to the learning curves. This suggests that learning from imperfect labelers is highly sensitive to the level of noise. Due to time constraints, we were unable to explore how best to quantify the level of noise, or to systematically explore the behaviour of baselines at different noise levels, and welcome further work. These preliminary results suggest that when learning from imperfect labelers in other contexts, a range of approaches should be tried and compared, and that the relative accuracy and bias of baselines will vary.

\end{document}